# UNet++ and LSTM combined approach for Breast Ultrasound Image Segmentation

Authors:

Saba Hesaraki1*, Morteza Akbari 2*, and Ramin Mousa3*

1*Faculty of Mechanics, Electrical Power and Computer, Islamic

Azad University Science and Research Branch, Tehran, Iran.

2* Department of Electrical and Computer Engineering, University of Birjand, Birjand, Iran

3* Department of Electrical and Computer Engineering, University of Zanjan, Iran

*Corresponding Author:

Saba.Hesaraki

**Abstract**
   Breast cancer stands as a prevalent cause of fatality among females on a global scale, with prompt detection playing a pivotal role in diminishing mortality rates. The utilization of ultrasound scans in the BUSI dataset for medical imagery pertaining to breast cancer has exhibited commendable segmentation outcomes through the application of UNet and UNet++ networks. Nevertheless, a notable drawback of these models resides in their inattention towards the temporal aspects embedded within the images. This research endeavors to enrich the UNet++ architecture by integrating LSTM layers and self-attention mechanisms to exploit temporal characteristics for segmentation purposes. Furthermore, the incorporation of a Multiscale Feature Extraction Module aims to grasp varied scale features within the UNet++. Through the amalgamation of our proposed methodology with data augmentation on the BUSI with GT dataset, an accuracy rate of 98.88%, specificity of 99.53%, precision of 95.34%, sensitivity of 91.20%, F1-score of 93.74, and Dice coefficient of 92.74% are achieved. These findings demonstrate competitiveness with cutting-edge techniques outlined in existing literature.

**Introduction**
Breast cancer is a condition that disrupts the process of cell division within the breast, leading to uncontrollable multiplication of cells beyond what is necessary. It is categorized into various types depending on its extent of spread and is recognized as a prevalent and significant malignant tumor among women globally[1]. The timely detection and treatment of this disease hold utmost importance. Typically originating in the fibroglandular region of breast tissue, early intervention plays a crucial role in reducing mortality rates associated with cancerous growths[2]. The impact of breast cancer is felt worldwide and encompasses both benign and malignant forms. Benign tumors pose minimal risks as they remain localized, whereas malignant tumors, also known as cancers, have the potential to spread to other body parts, posing a considerable threat. Early identification and classification of breast cancer are pivotal in preventing fatalities. Screening methods such as mammography and clinical breast examinations aid in the early detection of breast abnormalities[3]. While the WHO advocates for regular mammography screenings in developed nations with adequate resources and health awareness, this may not be viable in underdeveloped regions with limited resources. Educating individuals about breast cancer is crucial for early tumor detection[4]. Invasive ductal carcinoma (IDC) stands out as a predominant and critical form of breast cancer, characterized by the invasion of breast tissue[5]. Ultrasound imaging serves as a valuable tool in cancer diagnosis[6],[7] and visualization[3], offering a safe, real-time, and non-invasive method to assess internal body structures[8]. Despite its clinical utility, ultrasound imaging encounters challenges such as artifacts and noise, complicating both manual and automated analysis tasks[9].

These imperfections in images, such as stains and clutter, can pose challenges for basic thresholding and filtering algorithms, thus requiring the application of more advanced techniques to achieve accurate segmentation[10]. Historically, segmentation processes have been conducted after image acquisition, resulting in increased computational complexity. Analyzing MRI data manually is not only time-consuming and expensive but also vulnerable to human errors[11]. Despite the development of specialized systems for breast lesion detection claiming to improve diagnostic efficiency, many of these systems only target specific issues, leaving the automatic detection of breast lesions as an ongoing obstacle. Recently, there has been a notable focus on deep learning-based systems for cancer detection in medical imaging[9-16]. Breast cancer, known for its high fatality rates, demands early and precise detection methods, prompting the proposal of various techniques to enhance diagnostic precision. Among the array of medical imaging modalities, breast ultrasound stands out due to its cost-effectiveness and ability to accurately identify breast tumors in various breast regions. This study introduces a novel strategy for breast segmentation by fusing UNet++ and LSTM models. This hybrid model surpasses the conventional UNet and other standard methods by incorporating spatial and temporal feature extraction. Specifically, UNet++ is utilized for spatial feature capture, while LSTM is harnessed for temporal feature learning. Our research developed a pipeline for breast region segmentation integrating UNet++ and LSTM models, and evaluated its accuracy against traditional UNet and other baseline models, illustrating superior performance. The amalgamation of spatial and temporal features in our model notably enhanced segmentation accuracy, presenting a significant progression in the domain of breast cancer detection through ultrasound imaging, thus providing a robust approach to improve early diagnosis and treatment.

**Related Works**
Recent deep learning techniques, specifically convolutional neural networks (CNNs), have been extensively utilized in the field of medical image analysis[9]. Various versions of CNNs like SegNet and Unet have the capability to extract high-level features and conduct organ segmentation[10][11][12]. DGANet, introduced by researchers[13], is a dual global attention neural network designed for detecting breast lesions in ultrasound images, surpassing models such as YOLOv3 and Faster R-CNN. AMS-PAN[14] is an alternative segmentation method that integrates attention mechanisms and multi-scale features, achieving remarkable accuracy on BUSI and OASBUD datasets. Moreover, BTEC-Net[15] is a multi-stage approach for segmenting breast tumors by combining DenseNet121 and ResNet101 in an ensemble, outperforming traditional segmentation models like UNet on BUSI and UDIAT datasets. CSwin-PNet[16] is another effective strategy that employs a CNN-Swin Transformer coupled with a pyramid network for segmenting breast lesions in ultrasound images. This approach is based on datasets from the UDIAT Diagnostic Center of the Parc Tauli Corporation, Sabadell (Spain)[17] and Baheya Hospital, Cairo (Egypt)[18]. Evaluation against various approaches like UNet, AU-UNet, U-Net++, FPN, ViT, TransUNet, and Swin-UNet revealed a notable 10.36 improvement in IoU metric on dataset 1 and a 9.75 enhancement on Dataset 2. U-net++ is a method[19] proposed for Breast Regions Segmentation, incorporating encoder and decoder components linked by dense convolution blocks. It bridges the semantic gap between encoder and decoder feature maps before merging, with the encoder followed by a decoder on the left side. Unlike U-net, U-net++ features hopping paths connecting the encoder and decoder for deep monitoring. The approach was assessed using the DCE-MRI dataset containing 165 breast cancer images, achieving an IOU of 83.14 compared to U-net's IOU of 78.32. Additionally, Adaptive attention U-net (AAU-net)[20] has shown promising outcomes in comparison to UNet and UNet++. AAU-net consists of U-net with four down-sampling, four up-sampling, and four hop connections. Notably, the model incorporates a hybrid adaptive attention module (HAAM) in place of the original convolutional layer to enhance segmentation of breast lesions. Each encoding or decoding step in the model includes two HAAMs, with convolution layers of varying kernel sizes in HAAM offering diverse scale receiver fields to adapt to different input images. Furthermore, AAU-net can generate representations from BUS images through robust channel dimension and spatial dimension constraints, with the HAAM module comprising convolution layers with different kernel sizes, channel self-attention block, and spatial self-attention block.

The efficacy of this approach was assessed in three distinct datasets in the field of Business (BUSI), comprising a total of 780 images. Dataset B, on the other hand, contained 163 images, while the STU dataset consisted of 42 images. Furthermore, various alternative methodologies including AGNet, SANet, SENet, ECA-Net, and scSENet were scrutinized for the sake of comparison. The AAU-net methodology managed to attain a Dice coefficient of 77.51 on the BUSI dataset, whereas the most effective comparative model, namely ECA-Net, achieved a similar Dice coefficient of 77.51. ECA-Net demonstrated a Dice coefficient of 78.14 on Dataset B, whereas the superior competing method, AGNet, achieved a Dice coefficient of 73.30. Additionally, a myriad of other neural networks have been leveraged for the classification and segmentation of medical imagery, with notable mentions being the UNet Transformer [21] for Medical Image Segmentation, UNETR++ [22] and UNETR [23] for 3D Medical Image Segmentation, UNET 3+ [24] for Medical Image Segmentation utilizing Full-Scale Connected UNET Swin-Unet [25], employing an Unet-like Pure Transformer for Medical Image Segmentation, as well as ResUNet++ [26] and R2U-Net[27]. The comparative analysis of the various methodologies under investigation is delineated in Table 1.

**Method**
The objective of this article is to present a methodology aimed at enhancing the accuracy of extracting benign and malignant cancerous tumors within a network framework. Through the incorporation of multitasking attention and learning mechanisms into the traditional UNet++ architecture, an end-to-end encoder-decoder network specifically designed for automatic tube extraction has been developed (as illustrated in Figure 1). The proposed model consists of three primary components: the multiscale feature extraction module, the attention block, and the multitask learning module. Initially, the images undergo processing in the multiscale feature map extraction module to generate multiscale feature maps, which are then further refined to promote the amalgamation of multiscale features. Subsequently, the attention block is utilized to reinforce and capture the amalgamation of feature maps across various hierarchical levels. Lastly, the extracted feature maps undergo feature importance learning and time series feature extraction. The subsequent sections delve into the detailed explanation of the proposed methodology.

showcasing the investigated approaches along with their respective reference models, datasets used, evolution metrics, and identified weaknesses. For instance, the Segmentation-based Enhancement (SBE) approach failed to offer a comprehensive depiction of the model block and its parameters, while the Breast Tumor Ensemble Classification Network (BTEC-Net) exhibited challenges in delineating clear tumor borders in certain input images, resulting in irregular mask production. Conversely, the Pyramid Attention Network Combining Attention Mechanism and Multiple-Scale Features (AMS-PAN) model achieved lower IoU values in comparison to other UNet-based methodologies. Lastly, the UNet++ model applied in DCE-MRI solely evaluated performance based on IoU without considering additional evaluation metrics
.

**Table 1** Comparison of the investigated approaches.

| Reference | Model | Dataset | Evolution metrics | Weakness |
|---|---|---|---|---|
| [28] | Segmentation-based enhancement (SBE) | BUS<br>BUSI | IoU<br>Precision<br>Recall<br>F 1<br>mAP | Failure to provide a proper view of the model block and its parameters. Failure to make comparisons with approaches such as Unet |
| [13] | dual global attention neural network (DGANet) | BUSI | mAP | Failure to provide other metrics such as recall, precision and F 1 |
| [14] | Pyramid Attention Network Combining Attention Mechanism and Multiple- Scale Features (AMS-PAN) | BUSI<br>OASBUD | Accuracy<br>Precision<br>Recall<br>F 1<br>Specificity<br>Dice<br>IoU | It has obtained lower IoU values compared to other Unet- based approaches |
| [15] | Breast tumor ensemble classification network (BTEC-Net) | BUSI<br>UDIAT | Accuracy<br>Precision<br>Recall<br>F 1<br>Pixel accuracy<br>IoU<br>Dice | Due to the fact that the border of the tumor is unclear in some input images, the produced mask is in some cases irregular |
| [16] | CNN-Swin Transformer combined pyramid network | UDIAT<br>Baheya Hospital | Accuracy<br>Precision<br>Recall<br>F 1<br>IoU<br>Dice | The complexity of the model and the increase of learning parameters of the model compared to other approaches |
| [19] | UNet ++ | DCE- MRI | IoU | No other evaluation metrics were used |

**Multiscale Feature Extraction Module**

The use of encoder-decoder architectures such as Fully Connected Network and U-Net is widespread for extracting multi-scale features from images. These architectures can effectively combine deep feature maps from the decoder sub-network to enhance learning. UNet++ [26] is a variation of U-Net designed to bridge the information gap between encoder and decoder feature maps before merging. This study introduces the UNet++ multiscale feature extraction module and a proposed LSTM based on UNet++. In UNet++, the encoder and decoder subnets are linked through a series of nested and dense paths. Furthermore, long-range connections are established by the respective encoder and decoder sections. Consequently, the decoder section can integrate various hierarchical feature maps from the encoder, leading to enhanced accuracy and scalability of the network. In Figure 1, xi,j represents the output of node Xi,j, where i indicates the downsampling layer along the decoder path and j denotes the j-th convolutional layer along the hopping path. The feature maps xi,j can be represented as follows:

$$X_{i,j} = \begin{cases} \Psi^{Conv}(X^{(i-1,j)}) & j = 0 \\ \Psi Conv(\Psi cat(\Psi cat(x_{i,0}, x_{i,1}, ..., x_{i,j-1}), \Psi up(x_{i+1,j-1}))) & j > 0 \\ \Psi Conv(\Psi cat(\Psi cat(x_{i,0}, x_{i,1}, ..., x_{i,j-1}), \Psi up(x_{i+1,j-1}))) & j > 0 \end{cases}$$

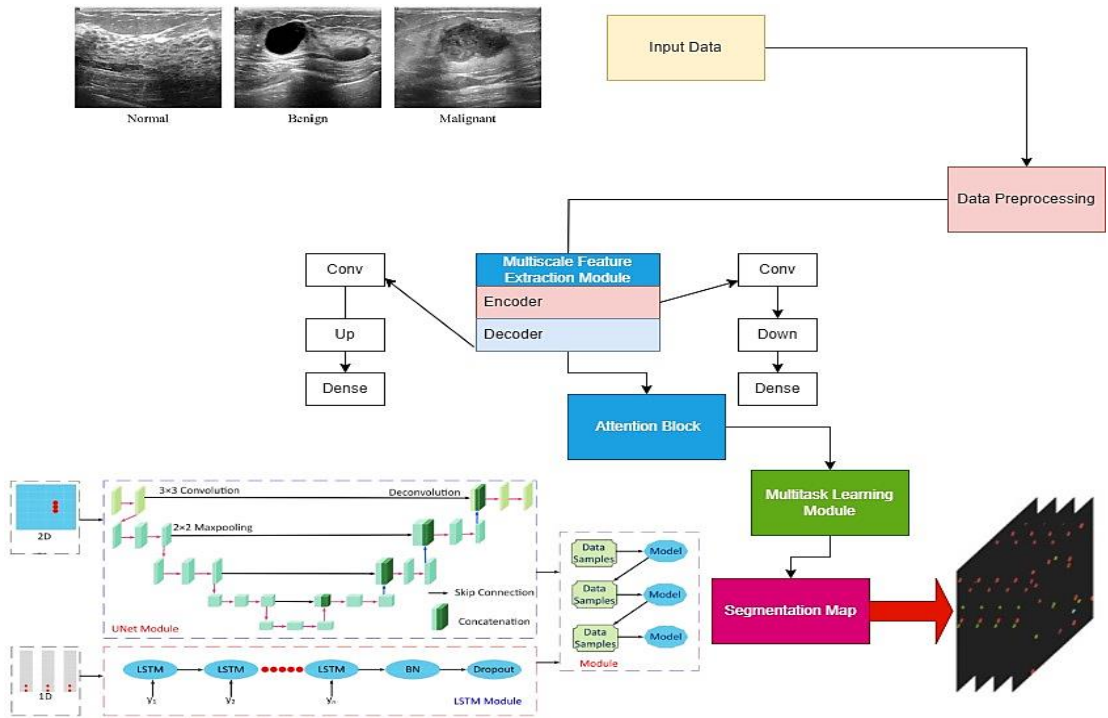

**Fig.1** Architecture of the Proposed UNet++LSTM

The symbol ΨConv denotes a convolution operation that has an activation function applied to it. The operation represented by Ψcat is a concatenation, while Ψup stands for an upsampling layer. When j = 0, Xi,j corresponds to the nodes in the encoder subnet. In cases where j > 0, Xi,j denotes the combined results of all other nodes at the same level, and the resultant sample contains deeper, coarser, and more semantic information.

Convolution entails applying a fixed-size kernel to the input matrix. At each step, the elements within the kernel are aggregated by multiplying the kernel matrix and the area in the input matrix that the kernel overlaps. Additional parameters may also be employed. The input matrix is padded with zeroes to maintain its size, and the Stride parameter determines how many elements to skip. An important aspect of the circular layer is the number of filters, which dictates the layer's depth. Each filter is trained to detect various image features within the input. The collector layer takes an input of size W1∗H1∗D1 and requires four parameters: the number of filters (K), kernel size (F), step (S), and zero pad (P). It generates an output layer of size W2 ∗ H2 ∗ D2, with the specified relationships holding true [37].

$$W_2 = \frac{(W_1 - F + 2P)}{S + 1}$$

$$H_2 = \frac{(H_1 - F + 2P)}{S + 1}$$

$$D_2 = K$$

The operation of convolution involves combining two functions to generate a third function and is typically applied to continuous functions. Convolution can be expressed by the following relationship [38]:

$$(I * K)(t) = \sum_{a \in D_I} I(a)k(t - a)$$

In the structure of deep learning, I represents input, and k represents the kernel. CNNs are commonly depicted as rectangular grids of real numbers, in which instance the convolution is two-dimensional and discrete. Typically, convolution is commutative, indicating that the kernels are moved over the inputs. Using these definitions, convolution can also be described as follows [39]:

$$(I * K)(i, j) = \sum_{1}^{m} \sum_{1}^{m} I(i - m, j - n)k(m, n)$$

In the equation I ∈ R(w ∗h), where I is the input, K ∈ R(m∗n) represents the kernel, and i and j are the coordinates for the output. It's important to mention that most modern deep-learning libraries utilize cross-correlation instead of convolution. The key distinction between cross-correlation and convolution is that cross-correlation doesn't involve flipping the kernels. Flipping entails rotating the kernel by 360 degrees. Hence, the kernel definition can be expressed as follows based on these assertions [39]:

$$(I * K)(i, j) = \sum_{1}^{m} \sum_{1}^{m} I(i + m, j + n)k(m, n)$$

**Attention Block**

In Figure 1, the four predicted feature maps produced by UNet++ are represented by X0,0, X0,1, X0,2, and X0,4. In UNet++, these feature maps are directly concatenated using an averaging operation, which results in ignoring the better feature maps and giving more importance to the negative feature maps in the final output. This neglect is primarily due to two reasons:

1. Low-level feature maps extract hierarchical semantic features with poor semantic information but rich spatial information (small receptive fields).

2. High-level feature maps have strong semantic information but weak spatial information because they possess large receptive fields.

Therefore, it is essential to distinctly combine feature maps from different levels so that the network allocates reasonable attention to both high-level and low-level features. Furthermore, due to the spatial impact of similar patterns and background noise on the extracted features, it is essential to highlight important parts and disregard unimportant ones. Hence, our approach involves assigning weights to prioritize the most crucial features when integrating feature maps from various levels. These weights are allocated to feature maps across different layers based on their relative significance in the channel dimension. Consequently, we have incorporated the Convolutional Block Attention Module (CBAM) into our proposed network. This module's purpose is to discern which information should be accentuated or suppressed. As depicted in Figure 2, the predicted map of $F \in R (C*H*W)$ is produced by the Multiscale features extraction module.

The primary objective of channel attention is to analyze significant features in an input image by leveraging the inter-channel connections of features. Initially, two operations, FC AVG and FC MAX, were developed to aggregate general information from each channel and discern distinct object features, respectively. These two descriptors are then input into an LSTM (shown in Figure 3) to generate two vectors, which are subsequently combined through element summation to yield the final channel attention map Mc (F).

$$M_c(F) = Sigmoid(lstm(lstm(F_{AVG}^C) + lstm(lstm(F_{MAX}^C)))$$

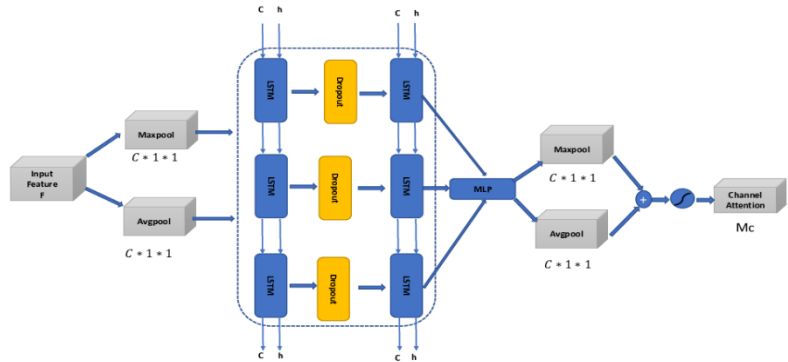

**Fig. 2** The proposed channel attention mechanism

The basic LSTM cell is composed of three gates: the forgetting gate (ft) decides what proportion of the previous data to forget. The input gate (it) assesses the information to store in the cell memory. Additionally, the output gate (ot) defines the approach for computing the output based on the current information.

$$ft = σ(W\ eif\ xt + U\ eif\ ht−1 + beif )$$

$$it = σ(W\ eiixt + U\ eiiht−1 + beii)$$

$$ot = σ(W\ eioxt + U\ eioht−1 + beio)$$

$$c_t = f_t \circ c_{t-1} + i_t \circ (W_{eic} x_t + U_{eic} h_{t-1} + b_{eic})$$
$$h_t = o_t \circ \tau(c_t)$$

The size of the input is denoted as 'n', while the size of the cell state and output is denoted as 'm'. The input vector at time 't' is represented by 'xt' (size n × 1), the forget gate vector by 'ft' (size m × 1), the input gate vector by 'it' (size m × 1), the output gate vector by 'Ot' (size m × 1), the output vector by 'ht' (size m × 1), and the cell state vector by 'ct' (size m × 1). The input gate weight matrices are represented as [Weif, Weii, Weio, Weic] (size m × n), while the output gate weight matrices are represented as [Ueif, Ueii, Ueio, Ueic] (size m × m). The bias vectors are denoted as [beif, beii, beio, beic] (size m × 1). The logistic sigmoid activation function is represented by σ, and the hyperbolic tangent activation function is represented by τ. The proposed approach for spatial attention examines areas of information that require more focus (refer to Figure 4). Pooling operations are used to collect information from individual channels and then combine them. Following this, a standard 7*7 convolutional layer is applied to the concatenated and convoluted channels to generate a spatial attention map.

$$M_s(F) = Sigmoid(Conv^{7*7}([F^S_{AVG}; F^S_{MAX}]))$$

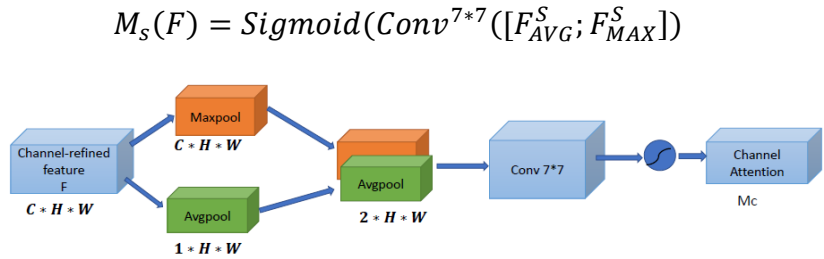

**Fig. 3** The proposed channel attention mechanism

The entire process of paying attention can be expressed in the following way:

$$F' = M_c(F) \otimes F$$

$$F'' = M_s(F') \otimes F'$$

The value of Mc in the real number set (C*1*1) represents the channel attention map, while Ms in the real number set (1*H*W) represents the spatial attention. The symbol "⊗" denotes element-wise multiplication, and F" represents the final feature map.

**Loss Function of proposed UNET++**
The proposed approach aims to reduce the pixel distance between tumor locations. In our plan, we will focus on minimizing this distance:

$$Dist(i) = \delta_d \min(\min_j d(i, j), d)$$

In an image, let i represent a pixel, x represent the group of pixels that make up tumor boundaries, and min minj denote the Euclidean distance between pixel i and the closest boundary pixel j. Let d be a threshold, and let δd represent a sign function that indicates whether the pixel is inside or outside the tumor. If δd = 1, the pixel i is within the tumor mask, whereas if δd = -1, it signifies that the pixel i is outside the tumor mask. Given these definitions of the multi-task loss, the proposed method can be described as follows:

$$L = \lambda_1 L_{seg} + \lambda_2 L_{DC}$$

The loss function for the segmentation task is denoted as Lseg, and the prediction of distance class maps is denoted as LDC. The learnable weights are represented by λ1 and λ2.
In simpler terms, the formula can be summarized in the following way:

$$L(x; \theta, \sigma_{DC}, \sigma_{seg}) = L_{seg}(x; \theta, \sigma_{seg}) + L_{DC}(x; \theta, \sigma_{DC})$$

The classification task's likelihood is determined based on the network output fc(x) at the end. This function is commonly known as softmax.

$$P(C = 1|x, \theta, \sigma_t) = \frac{\exp[\frac{1}{\sigma_t^2} f_c(x)]}{\sum_{c'=1} \exp[\frac{1}{\sigma_t^2} f_{c'}(x)]}$$

In this equation, P represents the predicted probability, fc(x) is the output obtained, fc′ (x) is the actual input, and σt is the Scaling factor. Broadly speaking, the classification error and uncertainty can be further expanded as follows:

$$(x; \theta, \sigma_t) = \sum_{c=1}^{C} -C_c \log P(C_c = 1|x; \theta, \sigma_t)$$

$$= \sum_{c}^{C} -C_c \log \left\{ \exp\left[\frac{1}{\sigma^2} f_c(x)\right] \right\} + \log \sum_{c'=1}^{C} \exp[\frac{1}{\sigma_t^2} f_{c'}(x)]$$

**Table 2** Settings of the proposed model and implementation environment

| Setting | |
|---|---|
| Operating System | Linux Ubuntu |
| Python | 3.9 |
| GPU | Tesla 4 |
| Image size | 128*128 |
| Batch size | 16 |
| optimizer | Adam |
| Learning Rate | 0.001 |
| Dropout rate | 0.5 |
| LSTM Block | 16 |

**Result**

The outcome of the study involved a comparative analysis of the proposed approach against 15 other established methodologies present in the existing literature. These methodologies include SD-CNN [28], CNN-GTD [29], GA-ANNs [30], SeResNet18[31], Faster R-CNN+CNNs [32], CNN+LR[33], ODET[34], Chowdary et al. [35], Inan et al.[36], Byra et al.[37], Shi et al.[38], and SaTransformer[39]. Furthermore, the assessment involved the evaluation of seven key criteria, namely accuracy, specificity, precision, sensitivity, F1-score, Jaccard, and dice. Similar to other methodologies documented in the literature, the proposed model allocated 80% of the data for training purposes and reserved 20% for model testing. The detailed configuration settings of the proposed model are outlined in table 2.

The outcomes derived from various methodologies are illustrated in table 3. The SD-CNN[28] methodology demonstrated the ability to achieve Accuracy=90.00, Specificity=94.00, and Sensitivity=83.00. Unfortunately, other assessment measures remain undisclosed for this specific methodology. In contrast, the CNN-GTD[29] methodology yielded inferior results compared to SD-CNN, with recorded values of Accuracy=86.50, Specificity=88.02, and Sensitivity=85.10. Regrettably, the remaining metrics remain undisclosed for this methodology. Surpassing the two aforementioned methodologies, the GA-ANNs [30] approach exhibited superior results. This particular approach achieved Accuracy=96.47, Specificity=95.94, and Sensitivity=90.00. The SeResNet1[31] methodology mirrored the performance of SD-CNN in terms of accuracy, attaining Accuracy=90.00. Additionally, this methodology achieved Precision=91.00, Sensitivity=90.00, F1-score=91.00, and Dice=90.00. The Faster RCNN+CNNs[32] methodology solely reported the Precision metric, which equated to 97.60, while other metrics remain undisclosed. The CNN+LR[33] approach demonstrated superior accuracy compared to the other methodologies under investigation, achieving Accuracy=06.87, Precision=86.93, F1-Score=82.64, and Dice=96.87. The ODET[34] methodology yielded elevated metric values and occupied a higher metric position than the methodologies under review. This specific approach achieved Accuracy=97.84, Specificity=98.63, Precision=93.28, Sensitivity=93.96, F1-score=93.16, and Dice=97.84. In contrast, the methodology proposed by Chowdary et al.[35] exhibited inferior outcomes in comparison to ODET, with results of Accuracy=88.08, Specificity=86.13, Precision=85.62, Jaccard=84.72, and Dice=84.81. Similarly, the approach introduced by Inan et al.[36] solely disclosed the Dice value of 63.40, indicating the weakest outcomes among all comparative methodologies. The approach implemented by Byra et al.[37] attained an accuracy of 92.33 on the specified dataset, alongside achieving Precision=81.00.

**Table 3** Result obtained by proposed UNet++LSTM and other comparative models available in the literature.

| Model | Accuracy | Specificity | PRE | Sensitivity | F1-score | Jaccard | Dice |
|---|---|---|---|---|---|---|---|
| SD-CNN [28] | 90.00 | 94.00 | - | 83.00 | - | - | - |
| CNN-GTD [29] | 86.50 | 88.02 | - | 85.10 | - | - | - |
| GA-ANNs [30] | 96.47 | 95.94 | - | 96.87 | - | - | - |
| SeResNet1 [31] | 90.00 | - | 91.00 | 90.00 | 91.00 | - | 90.00 |
| Faster RCNN +CNNs [32] | - | - | 87.60 | - | - | - | - |
| CNN+LR [33] | 96.87 | - | 86.93 | - | 82.64 | - | 96.87 |
| ODET [34] | 97.84 | 98.63 | 93.28 | 93.96 | 93.16 | - | 97.84 |
| Chowdary et al. [35] | 88.08 | 86.13 | 85.62 | - | - | 84.72 | 84.81 |
| Inan et al. [36] | - | - | - | - | - | - | 63.40 |
| Byra et al. [37] | 92.33 | - | 81.00 | - | - | - | 64.00 |
| Shi et al. [38] | - | - | - | - | - | 76.00 | 84.00 |
| Sa Transformer [39] | 93.34 | 88.32 | 89.51 | - | - | 83.12 | 86.334 |
| Unet++ | 98.58 | 99.56 | 94.53 | 87.17 | 90.70 | - | 90.70 |
| **Proposed model (UNet++LSTM)** | **98.88** | **99.53** | **95.34** | **91.20** | **93.74** | - | **92.74** |

In research by Shi et al.[38], a Jaccard similarity score of 76.00 and a Dice score of 84.00 were attained. Another approach, the SaTransformer approach[39], achieved an accuracy of 93.34, which was lower than most comparative methods. Furthermore, this approach yielded Specificity=89.51, Precision=89.51, sensitivity=89.51, Jaccard=83.12, and Dice=86.33.

Using tabularx and booktabs, the proposed Unet++ approach, combined with data augmentation, outperformed existing methods with an accuracy of 92.70. Additionally, it showed a 1.25 improvement in precision compared to the best comparative approach and performed better than ODET in terms of specificity. The proposed approach achieved accuracy=98.88, Specificity=99.53, Precision=95.34, sensitivity=91.20, F1-score=93.74, and Dice=92.74. Compared to Unet++, this approach showed an improvement of 0.0204 in Dice and 0.0304 in the F1-score metric.

Figures 4 and 5 show the performance diagrams of both the Unet and the proposed approach. Analyzing the Loss diagram reveals that neither the Unet nor the proposed approach exhibit underfitting, and their logarithmic learning form prevents overfitting. However, the proposed approach displays vibrations in the graph at higher epochs, which require adjustments. The BUSI dataset with GT1 contains medical images of breast cancer obtained from ultrasound scans. The data is categorized into normal, benign, and malignant tumors, and it comprises breast ultrasound images from women aged 25 to 75. This dataset was curated in 2018 and encompasses data from 600 female patients.

There are 780 images in the dataset, and the average image size is 500 x 500 pixels. The images are in PNG format, and there are corresponding ground truth images for each original image.

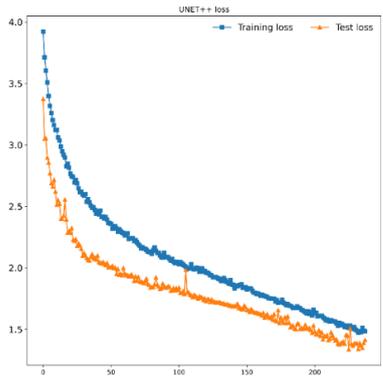

(a) Loss

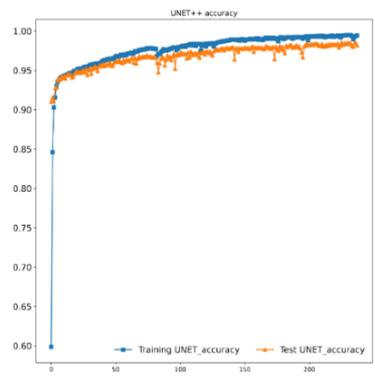

(b) Accuracy

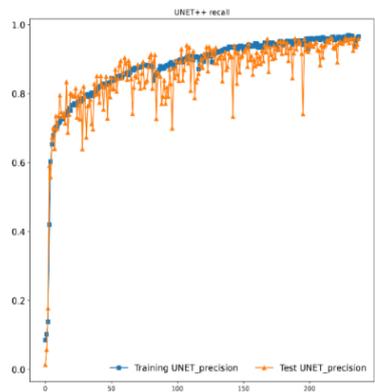

(c) Recall

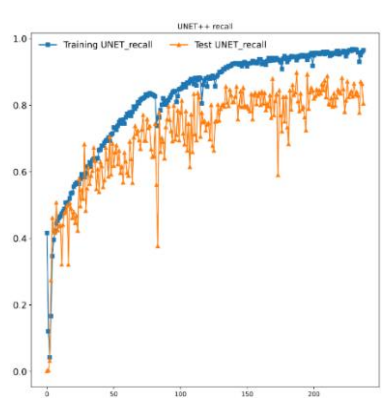

(d) Precision

**Fig.4**
Functional diagram of Unet++ approach, which does not have overfitting and underfitting

The model proposed in this study has demonstrated strong performance in commonly used segmentation measures, particularly the Dice criterion. The proposed model achieved Dice scores of 90.70 and 92.74 in two modes, Unet++ with data augmentation and the proposed model with data augmentation, indicating clearer images in BUSI dataset. The key area for enhancing the proposed approach lies in improving Accuracy, Precision, Specificity, and F1-secure metrics, particularly in relation to segmentation. Here are some recommendations for enhancing future work.

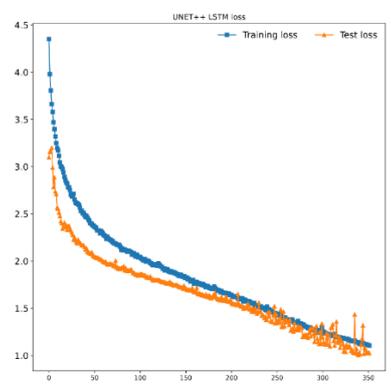

(a) Loss

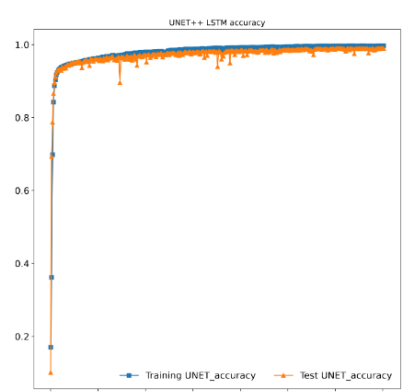

(b) Accuracy

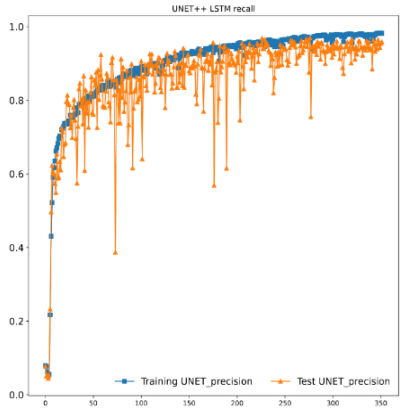
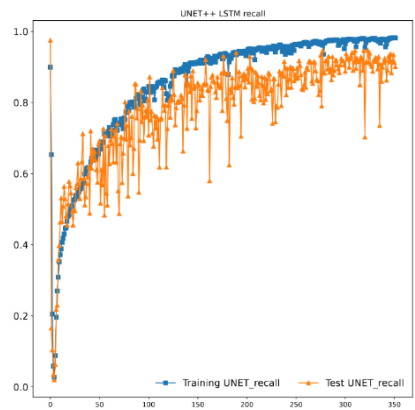

(c) Recall	(d) Precision

**Fig. 5** Functional diagram of Unet++LSTM approach, which does not have overfitting and underfitting

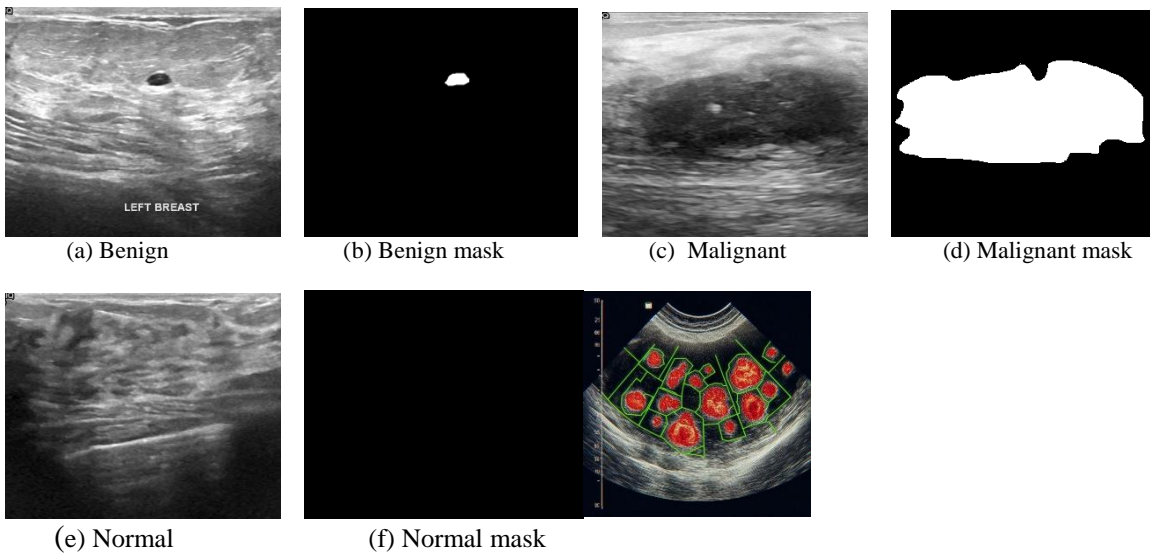

(a) Benign	(b) Benign mask	(c) Malignant	(d) Malignant mask

(e) Normal	(f) Normal mask

**Fig. 6** Samples of Dataset and image segmentation with proposed method

**Conclusion**

This research introduces UNet++LSTM, a new model designed for segmenting breast ultrasound images. The model incorporates a multiscale feature extraction module, an attention block, and a multitask learning module to effectively capture spatial and temporal features. When compared to 15 established models, our UNet++LSTM model demonstrated superior performance, achieving an accuracy of 98.88% and outperforming all other models.

The model achieved a specificity of 99.53%, with a precision of 95.34%, a sensitivity of 91.20%, an F1-score of 93.74%, and a Dice score of 92.74%, demonstrating significant advancements compared to existing techniques. The proposed method exhibited a 1.25% increase in precision compared to the top-performing comparative model and surpassed the ODET model in specificity. Both UNet++ and the proposed UNet++LSTM models demonstrated stable learning processes, as evidenced by the loss and accuracy diagrams, indicating avoidance of overfitting and underfitting. To further improve the model's performance and resilience, we suggest implementing transfer learning, exploring advanced parameter initialization techniques, and integrating rote learning strategies.

The BUSI dataset played a crucial role in training and testing our model, serving as a thorough evaluation platform to guarantee the credibility and applicability of our suggested method. In summary, the UNet++LSTM model shows significant progress in segmenting breast ultrasound images, providing an encouraging solution for the early and precise detection of breast cancer.

**Abbreviations**

**Acknowledgements**

**Authors' contributions**

**Funding**
The authors received no direct funding for this research.

**Availability of data and materials**

**Declarations**

**Competing interest**
The authors declare no competing interests.